\documentclass[sigconf]{acmart}

\AtBeginDocument{%
  }

\usepackage{hyperref}
\usepackage{balance}

\copyrightyear{2025}
\acmYear{2025}
\setcopyright{acmlicensed}
\acmConference[CogMAEC '25] {Proceedings of the 1st International Workshop on Cognition-oriented Multimodal Affective and Empathetic Computing}{October 27--31, 2025}{Dublin, Ireland.}
\acmBooktitle{Proceedings of the 1st International Workshop on Cognition-oriented Multimodal Affective and Empathetic Computing (CogMAEC '25), October 27--31, 2025, Dublin, Ireland}
\acmDOI{10.1145/3746277.3760413}
\acmISBN{979-8-4007-2059-8/2025/10}

\begin{document}

\title{Revisiting Your Memory: Reconstruction of Affect-Contextualized Memory via EEG-guided Audiovisual Generation}

\author{Joonwoo Kwon}
\authornote{ Four authors contributed equally to this research. Listing order is random.}
\affiliation{%
  \institution{Michigan State University}
  \city{MI}
  \country{USA}
}
\email{kwonjoon@msu.edu}

\author{Heehwan Wang}
\authornotemark[1]
\affiliation{%
  \institution{Seoul National University}
  \city{Seoul}
  \country{Republic of Korea}}
\email{dhkdgmlghks@snu.ac.kr}

\author{Jinwoo Lee}
\authornotemark[1]
\affiliation{%
  \institution{Seoul National University}
  \city{Seoul}
  \country{Republic of Korea}
  }
\email{adem1997@snu.ac.kr}

\author{Sooyoung Kim}
\authornotemark[1]
\affiliation{%
  \institution{Rutgers University}
  \city{NJ}
  \country{USA}
  }
\email{sooyoung.k@rutgers.edu}

\author{Shinjae Yoo}
\affiliation{%
  \institution{Brookhaven National Laboratory}
  \city{NY}
  \country{USA}
  }
\email{sjyoo@bnl.gov}

\author{Yuewei Lin\texorpdfstring{\textsuperscript{\dag}}{†}}
\affiliation{%
  \institution{Brookhaven National Laboratory}
  \city{NY}
  \country{USA}
  }
\email{ywlin@bnl.gov}

\author{Jiook Cha}
\authornote{Co-corresponding authors.}
\affiliation{%
  \institution{Seoul National University}
  \city{Seoul}
  \country{Republic of Korea}}
\email{connectome@snu.ac.kr}

\renewcommand{\shortauthors}{Kwon et al.}

\begin{abstract}
In this paper, we introduce \textbf{RevisitAffectiveMemory}, a novel task designed to reconstruct autobiographical memories through audio-visual generation guided by affect extracted from electroencephalogram (EEG) signals. To support this pioneering task, we present the \textbf{EEG-AffectiveMemory} dataset, which encompasses textual descriptions, visuals, music, and EEG recordings collected during memory recall from nine participants. Furthermore, we propose \textbf{RYM} (\textbf{R}evisit \textbf{Y}our \textbf{M}emory), a three-stage framework for generating synchronized audio-visual contents while maintaining dynamic personal memory affect trajectories. Experimental results demonstrate our method successfully decodes individual affect dynamics trajectories from neural signals during memory recall (${F1}=0.9$). Also, our approach faithfully reconstructs affect-contextualized audio-visual memory across all subjects, both qualitatively and quantitatively, with participants reporting strong affective concordance between their recalled memories and the generated content. Especially, contents generated from subject-reported affect dynamics showed higher correlation with participants' reported affect dynamics trajectories (${\mathbf r}=0.265$, ${\textit{p}}<.05$) and received stronger user preference (${preference}=56\%$) compared to those generated from randomly reordered affect dynamics. Our approaches advance affect decoding research and its practical applications in personalized media creation via neural-based affect comprehension. Codes and the dataset are available at \hyperlink{https://github.com/ioahKwon/Revisiting-Your-Memory}{https://github.com/ioahKwon/Revisiting-Your-Memory}.

\end{abstract}

\ccsdesc[500]{Computing methodologies~Artificial intelligence}
\ccsdesc[500]{Computing methodologies~Computer vision}
\ccsdesc[500]{Computing methodologies~Reconstruction}
\ccsdesc[500]{Human-centered computing~Human computer interaction (HCI)}
\ccsdesc[500]{Human-centered computing~User studies}

\keywords{Affective Computing; EEG-based Affect Decoding; Affect Decoding Dataset; Autobiographical Memory Reconstruction; Multimodal Generation; Personalized Media Generation; Neural Signal Interpretation; Audiovisual Synthesis; Multimodal Affective Computing; Affect-EEG }

\begin{teaserfigure}
  \includegraphics[width=\textwidth]{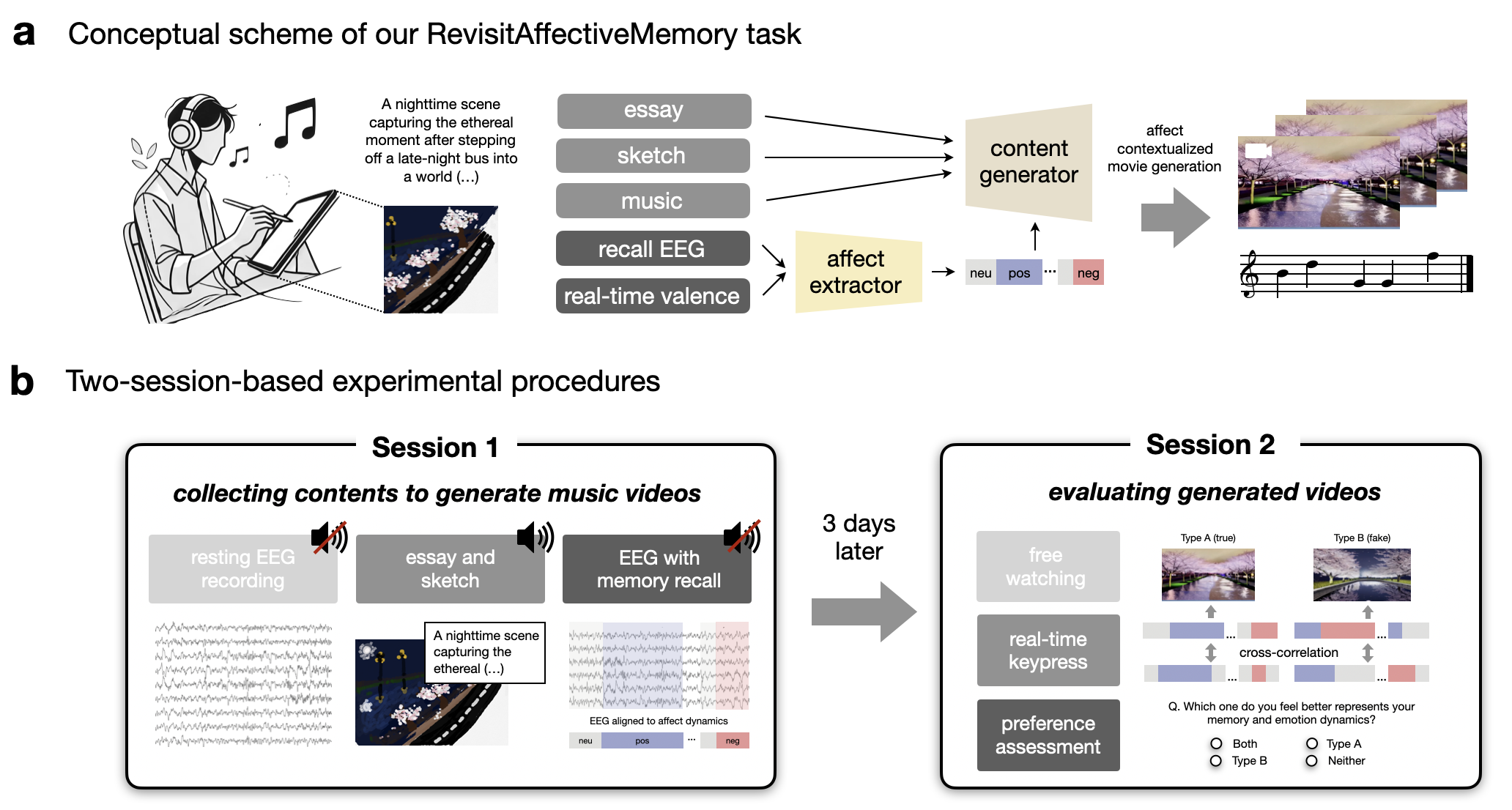}
  \caption{Study overview. (a) Conceptual scheme of \textbf{RevisitAffectiveMemory} task. (b) The \textbf{EEG-AffectiveMemory} dataset collection process with nine participants recalling their memories.}
  \label{fig:fig1}
  \Description{.}
\end{teaserfigure}

\received{30 June 2025}
\received[accepted]{5 August 2025}

\maketitle

\section{Introduction}

Autobiographical memories, constructed from sensory impressions and affective states, are central to shaping our sense of self. 
When familiar sights or sounds activate these memories, they bridge past and present experiences, allowing us to relive both sensations and affects.

Affective memory recall is a core area of research in cognitive and affective neuroscience, providing insights into how affect dynamics shape the richness and retrieval of personal experiences. Research has established that affect plays a crucial role in forming memories composed of visual and auditory information \citep{jancke2008music}. Autobiographical memory recall, in particular, may effectively evoke robust affective states, underscoring the deep interconnection between affect and memory \citep{siedlecka2019experimental}. Despite the co-occurrence of various affects during memory recall, prior studies have mainly focused on static self-reported measures, offering limited insights into the temporal dynamics of affect \citep{mills2014validity}. Recent research has shifted toward examining how multiple affective states change and interact over time in naturalistic scenarios \citep{kuppens2017emotion}. This approach employs stimuli such as movies or music to investigate both the neural substrates of affective states and their influence on sensory processing and memory formation \citep{mcclay2023dynamic}. 

This evolving focus has advanced affect decoding research, with studies demonstrating the successful application of machine learning for decoding affective states from neural signals, including movie viewing, music listening, and memory recall \citep{saarimaki2021naturalistic,chanel2009short,iacoviello2015real,dar2024insights}. While EEG-based affect decoding shows promise for real-world applications, its untapped potential in decoding the rich and dynamic affective states evoked during autobiographical memory recall remains a critical and largely unexplored frontier. Understanding these dynamics and developing methods for their neural decoding could lead to breakthroughs in uncovering the fundamental role of recalling unforgettable memories in everyday affective experiences.

In parallel, affect-guided content generation has rapidly evolved, opening new possibilities for transforming affect-contextualized memories into multimedia experiences. Early studies demonstrated basic affect-guided neural style transfer in images \citep{huang2018automatic} and music \citep{sigel2021music}, while recent advances have enabled more nuanced control of affective expression through affect-specific text prompts \citep{weng2023affective,agostinelli2023musiclm,copet2024simple}. Furthermore, recent generative models support more sophisticated control over the generation process \citep{rombach2022high}. 

Building on these advancements, the integration of generative models with affect decoding offers a novel opportunity to create multimedia content that preserves both the content and affective context of autobiographical memories. Given the critical role of affective states in shaping sensory processing during memory formation and recall, this integration has the potential to transform personalized entertainment and media creation, providing individuals with a deeply resonant experience where their autobiographical memories are enriched by AI-generated content.


Here, we aim to decode temporal dynamics of affective states from EEG during autobiographical memory recall and develop a system for generating affectively resonant videos accompanied by music from personal memories using generative artificial intelligence. To support this task, we present \textbf{EEG-AffectiveMemory} dataset, which captures multi-modal data including text descriptions, sketch paintings, associated musical pieces, and EEG signals during memory recall. Of note, our experimental protocol specifically targets affective state transitions by having participants recall episodes containing mixed affects (e.g., "\textit{sad but cozy}" and "\textit{happy but lonely}") and report real-time affective states during memory recall and generated content viewing. This experimental design enables simultaneous tracking of dynamic affective changes alongside their neural correlates.

Leveraging this comprehensive dataset, we propose \textbf{RYM} (\textbf{R}evisit \textbf{Y}our \textbf{M}emory), a three-stage multi-modal framework that decodes dynamic affective transitions from neural signals and generates synchronized audio-visual content reflecting the affective trajectories during memory recall. 
To summarize, our work has three main contributions.
\begin{itemize}
  \item For the first time, we introduce a novel task and comprehensive multi-modal dataset for studying affect-contextualized autobiographical memories. This dataset is accessible on our project page.
  \item We successfully decode dynamic affective states from EEG signals during memory recall, capturing temporal affective trajectories.
  \item We propose RYM, a simple yet effective framework that faithfully generates synchronized audio-visual content while preserving the affective dynamics of personal memories.
\end{itemize}

\begin{figure*}[t]
    \centering
    \includegraphics[width=0.85\textwidth]{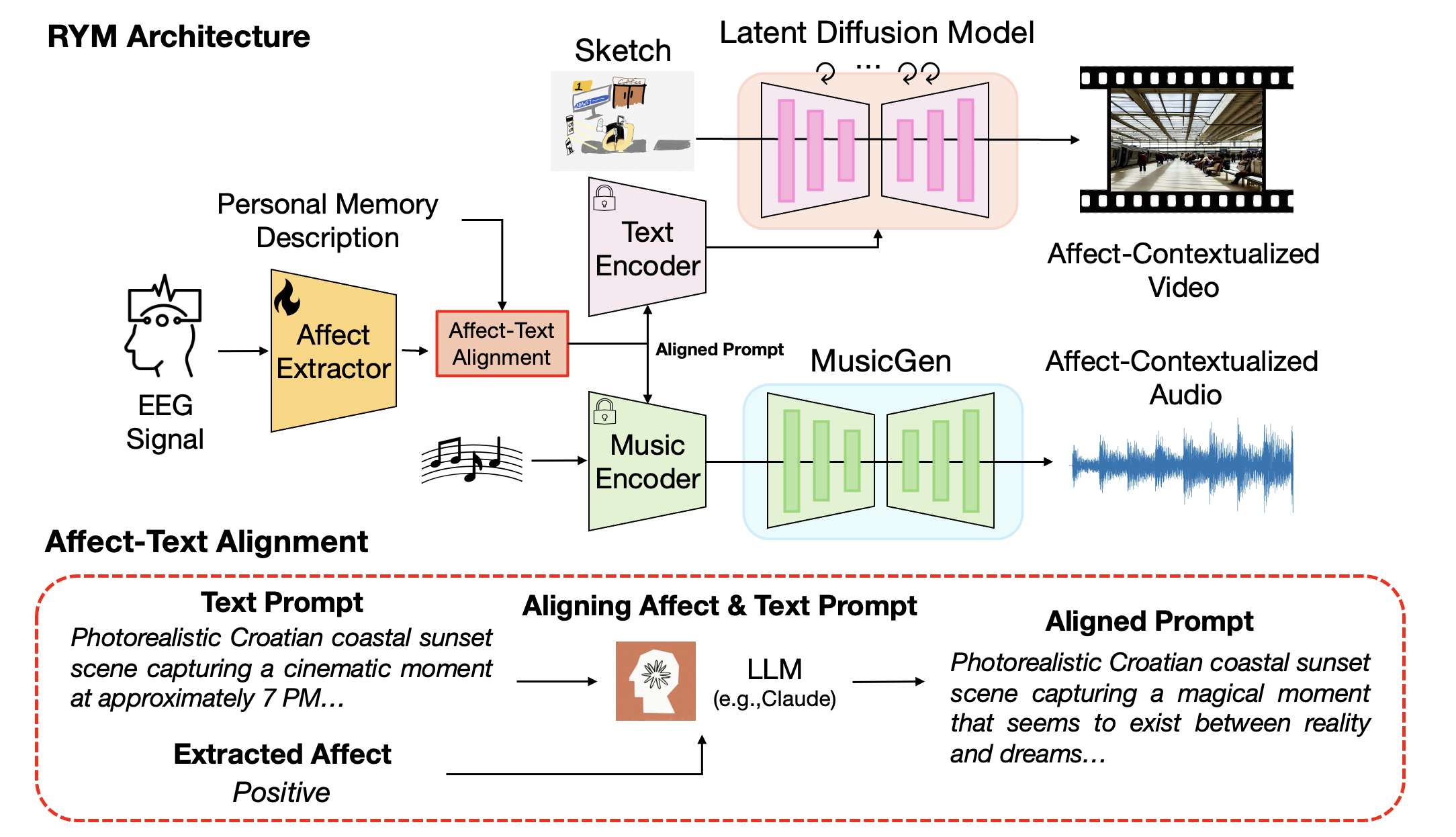}
    \caption{The \textbf{RYM} architecture comprises three primary components: an affect extractor, an affect-text alignment, and affect-contextualized decoding. The recorded EEG signals are fed to an affect extractor, making an dynamic affect representations, which are subsequently aligned with text prompts using pre-trained language models (e.g., Claude). A synchronized affect-contextualized text prompt is then processed into the music and visual generation models.}
    \label{fig:fig2}
    \Description{.}
\end{figure*}

\section{Related Works}

\subsection{Affect Decoding using Human Brain Signal} Machine learning models have shown increasing success in decoding affect from neural signals (EEG, fMRI) \citep{saarimaki2021naturalistic}. While traditional studies used controlled stimuli, research has shifted toward naturalistic approaches using movies and music \citep{koelstra2011deap,zheng2015investigating}. EEG has emerged as a preferred modality due to its high temporal resolution and accessibility \citep{al2017review,vempati2023systematic}. Several influential open-source datasets (DEAP, SEED, DREAMER) combine EEG with naturalistic stimuli, with DREAMER demonstrating portable EEG feasibility \citep{koelstra2011deap,zheng2015investigating,katsigiannis2017dreamer}. Recent datasets explore increasingly natural contexts: social video watching (AMIGOS) \citep{miranda2018amigos} and facial expression recording (PME4) \citep{chen2022emotion}. These resources have advanced both neural pattern analysis and affect dynamics decoding using machine learning and deep learning models \citep{saarimaki2021naturalistic,vempati2023systematic}. Despite these advances, the neural dynamics of autobiographical memory recall, a powerful affect elicitor, remain largely unexplored \citep{siedlecka2019experimental}. While early studies showed potential in decoding memory-induced affect using conventional machine learning \citep{chanel2009short}, subsequent research demonstrated successful affect classification from EEG signals \citep{iacoviello2015real}. Recent work shows deep learning models outperform conventional machine learning approaches in memory-induced affect classification \citep{dar2024insights}, while their requirements on larger training datasets significantly hinder their applications \citep{al2017review,vempati2023systematic}.
Recently, CEBRA \citep{schneider2023learnable} addresses the sample size limitation in affect decoding through contrastive learning, using temporally aligned variables like affect time-series. Demonstrating efficacy in various neural decoding tasks \citep{merk2023invasive,yi2024awe}, CEBRA significantly outperforms traditional approaches like frontal alpha asymmetry in cross-participant affect valence decoding \citep{yi2024awe}. Thus, we utilized CEBRA to effectively extract affect dynamics from individual subjects within a single session of neural recordings.

\subsection{Neural Style Transfer} 
Neural Style Transfer (NST) enables example-guided style transfer while preserving content \citep{gatys2016image,johnson2016perceptual,chen2016fast,ulyanov2016texture}.
The introduction of adaptive instance normalization \citep{huang2017arbitrary} and its extensions \citep{kotovenko2019content,chandran2021adaptive,wang2023microast} advanced arbitrary style transfer capabilities, though requiring intensive computation with pre-trained convolutional neural networks (CNNs). Recently, AesFA achieved lightweight implementations without pre-trained CNNs, enabling real-time high-resolution applications \citep{kwon2024aesfa}. Meantime, diffusion models have advanced style transfer through textual inversion \citep{zhang2023inversion} and CLIP-based disentanglement \citep{wang2023stylediffusion} to text-conditioned approaches \citep{everaert2023diffusion,kwon2022clipstyler,chung2024style}. Several works have explored affect-based style transfer using affect-color palettes \citep{huang2018automatic} and visually-abstract affects from text \citep{weng2023affective}, but remains limited to static representations. We address this limitation by proposing a framework that leverages dynamic affective information from EEG signals to generate affect-contextualized music videos.

\subsection{Music Style Transfer} 
Following NST in computer vision, music style transfer enables content and style separation in musical pieces \citep{dai2018music}. While lacking formal definitions, research frameworks typically treat melody, harmony, and rhythm as content, while considering timbre, articulation, and dynamics as style elements \citep{dai2018music,grinstein2018audio,cifka2019supervised}. Early music style transfer focused on timbre transformation while preserving structural content using WaveNet and adversarial networks \citep{engel2017neural,huang2018timbretron,grinstein2018audio}. Research evolved to broader style elements, enabling genre transformation through manipulation of melody, harmony, and instruments \citep{cifka2019supervised}. Recent work explores the affective expression of music, separating low-level features (pitch, harmony) from high-level features (rhythm, dynamics, tempo) for controllable style transfer \citep{{tan2020music}}. Recent advances in LLMs enable text/image-conditioned and melody-guided music generation \cite{agostinelli2023musiclm,copet2024simple,kim2024training}, simplifying synthesis through text/image descriptions and audio references. Latest approaches \cite{liu2025javisdit, sun2024mm, ruan2023mm} utilize multi-modal diffusion models for synchronized audio-visual generation, achieving strong alignment between generated modalities. However, none of these approaches directly utilize neural signals to extract affect, nor do they utilize affect dynamics. 


\section{Experimental Designs}
The experiment consisted of two sessions. 
In session 1, EEG signals and real-time affective ratings were recorded during memory recall. 
Participants also provided a brief essay and sketch of their recalled memories, which were later used to generate personalized music videos. 
Session 2 was conducted three days later to assess the generated music videos through a user study. 
A total of 10 participants were recruited and 9 participants who completed both sessions were included in the analysis (8 females; $\textit{M}_{age}$ = 24.1 years, $\textit{SD}_{age}$ = 1.5 years). 
All participants provided written informed consent before the experiments. All procedures were approved by the Institutional Review Board of Seoul National University. The experiments were carried out in a soundproof room.

\subsection{Experiment Session 1} 
One day prior, participants were instructed to prepare an affective memory, evoking mixed effects (for example, recalling a high school graduation may involve both sadness about leaving friends and excitement about college life), and a song that enhanced the vividness of their recall. The selected song had to have a piano solo cover available on a streaming platform.

The first session comprised four stages. 
First, participants were fitted with Enobio 20 EEG devices (Neuroelectrics), and signal quality calibration with 2-minute eyes-closed resting-state EEG recording was performed. 
Second, participants repeatedly listened to their chosen song while writing a 50-75 word essay in Korean describing their memory. 
Third, while listening to the same song, participants created a digital sketch of their memory using \textit{Tayasui Sketches} on an iPad. 
Last, during the memory recall, participants closed their eyes as EEG data was recorded and indicated affective changes using real-time keypresses. 
They kept pressing ‘1’ for positive, ‘3’ for negative, and no key for neutral states. 
Keypress events were temporally mapped to the EEG signals. 
Participants stopped recalling when no new scenes appeared or affective responses subsided. 
Afterward, participants rated their confidence in their keypress responses on a 1-7 Likert scale (i.e., 1 = Very insecure, 7 = Very confident) and reported acceptable performance ($mean$ = 5.2, $std$ = 1.0).
We used separate key presses instead of continuous rating because rating feelings constantly in real time with eyes closed is hard and could disturb the natural memory process \cite{jolly2022recovering}. Furthermore, several studies showed that similar key press methods for rating feelings in real time have worked well \cite{vaccaro2024neural, yi2024awe}.

\subsection{Experiment Session 2} Session 2 consisted of three stages.
To rigorously evaluate whether the contents generated with \textbf{RYM} properly reflect subjective affect dynamics, we compared "real" (based on their CEBRA-decoded affective sequence) against "fake" (based on permuted affective sequences) music videos. "Real" music videos were generated based on their CEBRA-decoded affect dynamics trajectories, while "fake" music videos used randomly reordered affect dynamics trajectories.
First, participants were told that both real and fake videos were created differently but authentically, and they freely watched both a "real" and a "fake" music video to reduce the confounding effects of surprise reactions during the following evaluation. 
Second, after a 30-second break, the participants re-watched each video and evaluated the affective changes depicted in the video using the same keypress method as in session 1. 
While session 1 keypresses reported subjective feelings, session 2 keypresses assessed affective recognition of the videos. 
Last, after another 30-second break, participants watched the videos again and rated which better represented their memory and its affective dynamics. 
They chose from four options: \textit{Both, Real, Fake, Neither}. 
The viewing order during keypress evaluation was randomized to reduce bias and stop participants from easily identifying which video matched the real sequence of feelings they remembered.

\begin{figure*}[t]
    \centering
    \includegraphics[width=\textwidth, height=11cm]{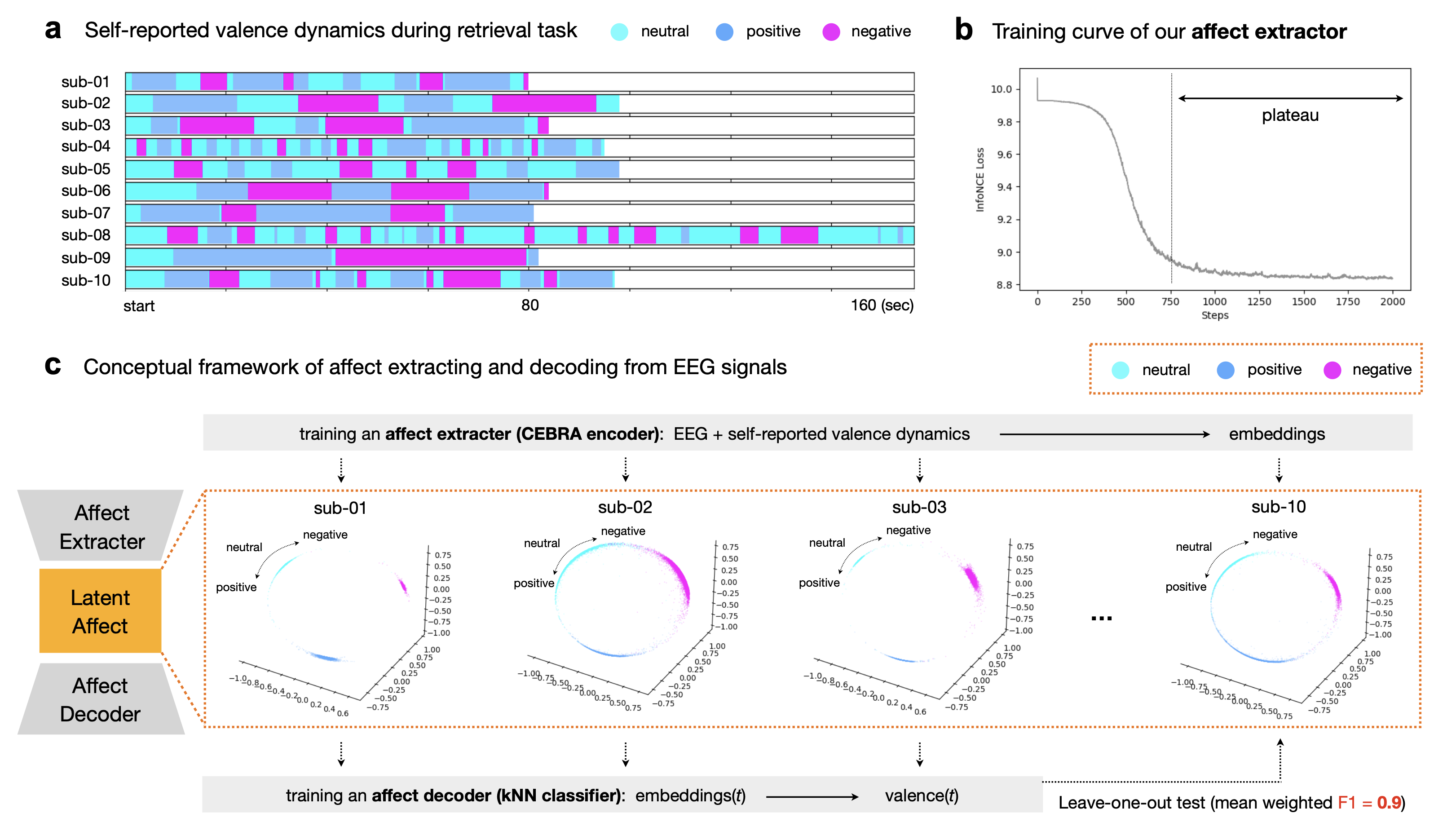}
    \caption{Our affect decoding results. (a) idiosyncratic temporal dynamics of real-time valence keypress during memory recall task. (b) learning curve of CEBRA's multi-session training. (c) each participant's latent neural representation of valence.}
    \label{fig:fig3}
    \Description{.}
\end{figure*}

\begin{figure*}[!t]
    \centering
    \includegraphics[width=\textwidth]{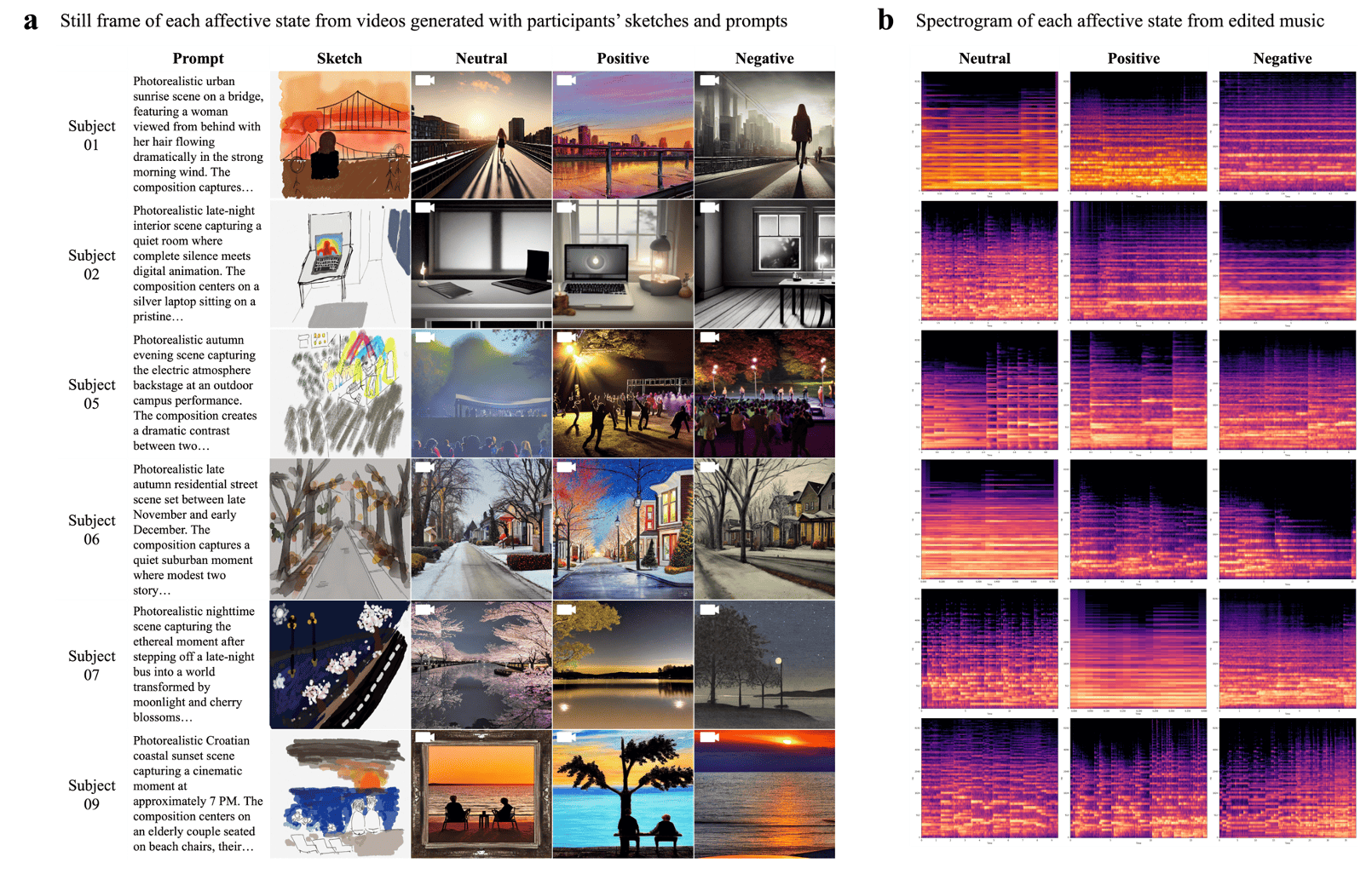}
    \caption{Qualitative results. (a) the images and (b) the music (mel-spectogram for visualization) are randomly selected. Note that the images are randomly sampled from the generated video output.}
    \label{fig:fig4}
    \Description{.}
\end{figure*}

\section{Method}
Here, we introduce a novel methodology that successfully transfers one's affect and/or preferences into desired audio/visual medium. 
Specifically, we utilized contrastive learning-based neural network to extract affective states from individual subjects' neural recordings during autobiographical memory recall. 
In addition, we explicitly aligned dynamic affective transitions with corresponding text descriptions using the pre-trained language model, which are then fed to the pre-trained generative models for music and video.
\subsection{Architecture Overview}
Depicted in Figure~\ref{fig:fig2}, the RYM architecture comprises three primary components: an affect extractor, an affect-text alignment, and affect-contextualized decoding. Specifically, the measured EEG signals are fed to an affect extractor, making dynamic affect representations which are then aligned with text prompt using pre-trained language models (e.g., Claude 3.5 Sonnet). An aligned affect-contextualized text prompt is then processed into the music and image generation models.

\subsection{Extracting Human Affects}
To effectively extract affects from individual subjects within a single session of neural recordings, we devised an 'affect extractor' based on CEBRA \cite{schneider2023learnable}, a deep neural encoder grounded in contrastive learning. By leveraging temporally synchronized behavioral sequences as auxiliary variables, CEBRA non-linearly reduces the dimensions of neural signals to maximize the separation between distinct behavioral states while clustering signals from similar states. Initially validated in human invasive brain-computer interfaces (BCI) applications \cite{merk2023invasive}, CEBRA has recently demonstrated superior performance in human affect decoding during movie viewing \citep{yi2024awe}, suggesting its potential for EEG-based affect decoding.

Our affect extractor utilizes keypressed valence sequences (neutral, positive, negative) as auxiliary variables to identify individual-level latent clusters for each affective state. We performed multi-session training with 10 participants' EEG and key-pressed valence sequence, aligning individuals' latent valence-neural representations. To examine the generalizability of embeddings, we conducted leave-one-out valence decoding tasks. We trained a k-nearest neighbor (KNN) classifier with 9 participants' identified embeddings to predict valence label at each timepoint and evaluated its predictive performance with the other's embeddings. 

\subsection{Aligning Extracted Affects With Text}
To effectively incorporate extracted affects into audio/visual output, it is essential to convert them into text prompts, which are commonly used for directing generative processes. Specifically, we created a word bank of 16 positive and 16 negative emotion words commonly used by our participant group for strong feelings based on \citeauthor{park2005making}, \citeyear{park2005making} \cite{park2005making} as follows: (positive)  \textit{peaceful, comfortable, refreshed, unburdened, energized, proud, enchanted, satisfied, jubilant, thrilled, joyful, worthwhile, funny, lovely, welcoming, touched}; (negative) \textit{overwhelmed, disappointed, miserable, resentful, lonely, betrayed, hateful, melancholic, restless, anxious, troubled, aggrieved, annoyed, frustrated, guilty, tedious}. For content generation, we randomly selected words from our word bank and used them constructing prompts. For videos, we fine-tuned participants' original memory descriptions with selected words using the pre-trained language model, \textit{claude 3.5 sonnet}, leaving neutral state prompts unaltered. For music, we used the template \textit{"A \{selected word\} song with only grand piano solo play. No other instruments play."} (omitting affective words for neutral states). These prompts, along with participants' musical pieces heard during experiments, were fed to MusicGen-melody for affect-contextualized.

\subsection{Implementation Details}
\textbf{Affect Extraction.} To decode each individual's affect dynamics from memory recall EEG signals, we conducted multi-session training of CEBRA. We set hyperparameters as follows: batch size=2,048, model architecture=‘offset-10 model’, the number of hidden units=95, learning rate=.005, the number of iterations=2000, hybrid=False. 
Following the previous study on the optimal number of dimensionality for generalizable valence representation across individuals and stimuli \citep{yi2024awe}, we set the number of latent embeddings' dimensionality as 7. 
The neighborhood parameters of KNN classifiers were fixed at the square of the number of the input time points.

\textbf{Music and Video Generation.} Using MusicGen-melody \citep{copet2024simple}, we generated music using affect-specific text prompts and memory-associated guiding melodies selected by participants. The generated content was then segmented according to affective state durations and integrated into a final sample using 40 ms crossfading to match the affective trajectory. For video generation, we used stable diffusion ver.1.5 \citep{rombach2022high}. In all our experiments, we fix the parameters of text encoders and LDMs (Latent Diffusion Models) for music and video generation, respectively. We use the author-released codes using default configurations. All experiments were conducted using the PyTorch framework \cite{paszke2019pytorch} on a single NVIDIA A100(40G) GPU.

\begin{figure*}[!t]
    \centering
    \includegraphics[width=\textwidth]{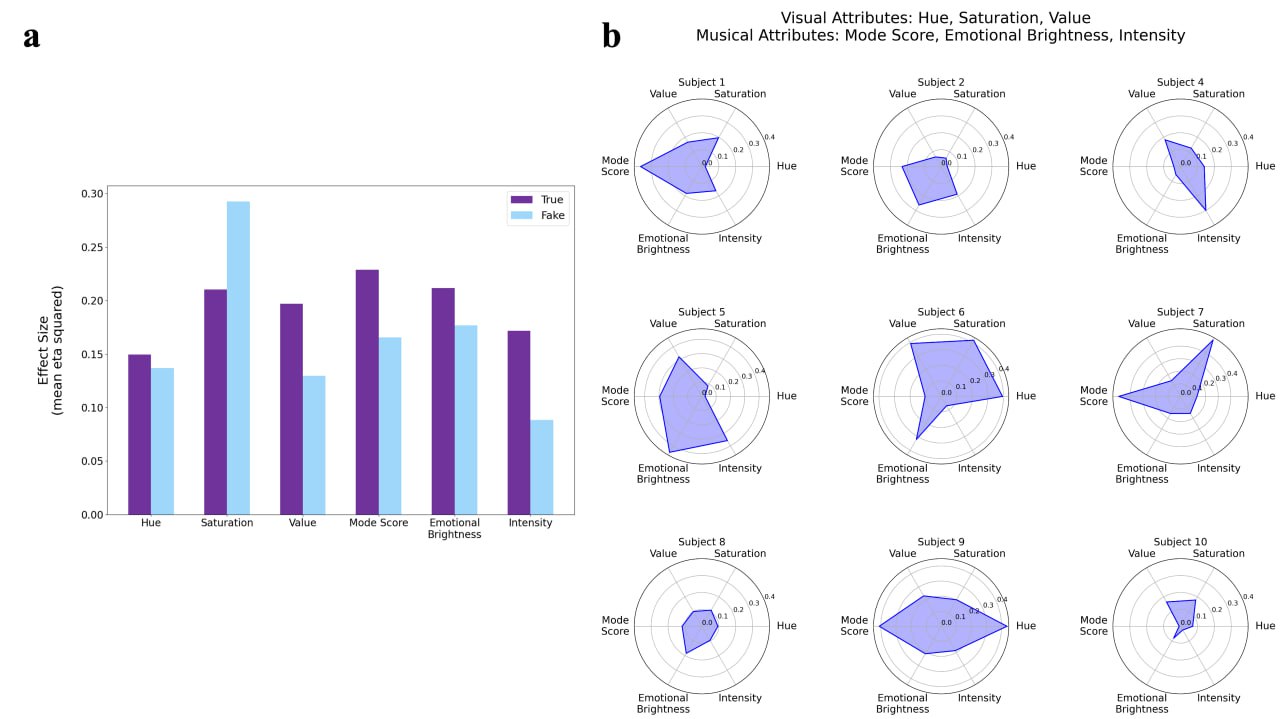}
    \caption{The affective correlations of visual and audio attributes. (a) Comparison of true and fake (i.e., random) music videos' visual/audio attributes in the difference between each affective state. (b) Individual differences in visual/audio attributes across each affective state. }
    \label{fig:fig5}
    \Description{.}
\end{figure*}

\section{Experimental Results}
To evaluate the performance of our models on the new task, we utilize metrics in three aspects: qualitative, quantitative and user-level.

\subsection{Affect Decoding} 
Participants showed individual differences in temporal dynamics of affective states during memory recall and the duration of the recall (see Figure~\ref{fig:fig3}a). Despite these variations, our affect extractor demonstrated a plateau in learning performance in multiple-session training with EEG signals and real-time valence keypresses from 9 participants (see Figure~\ref{fig:fig3}b). The individually derived latent valence-neural embeddings from this model revealed a consistent geometric structure across individuals (see Figure~\ref{fig:fig3}c). Using these embeddings with a KNN classifier for leave-one-out classification for each individual yielded a test-weighted F1 score of 0.9.

\subsection{Qualitative Evaluation} 
Figure~\ref{fig:fig4} illustrates that our approach produces qualitatively satisfactory visual and audio outputs across all affective states and subjects. 
Our model preserves the elements of the original sketch while integrating corresponding affects through tones, colors, and visual composition. 
The positive video clip showed brighter colors and a vibrant tone throughout the videos, whereas the negative displayed darker tones and a gloomier atmosphere. 
Moreover, the generated output demonstrated satisfactory visual coherence across video frames. 
Also, our model successfully transfers the style of original guiding melodies. 
As in Figure~\ref{fig:fig4}, the positive affects reflecting music clips showed broader frequency distributions with stronger energy in high-frequency regions, creating brighter and more vibrant spectral patterns. 
In contrast, the negative affects reflecting content showed concentrated energy in lower frequency bands, resulting in darker and more focused spectral patterns. 
These results demonstrated clear acoustic differences between affective expressions in generated music.
The down-sampled version of generated music samples can be accessed within the supplementary materials.

\begin{table}[ht]
\renewcommand*{\arraystretch}{1.2}
\centering \resizebox{0.45\textwidth}{!}{
\begin{tabular}{cc|cc}
             & Output  & CLIP($\downarrow$) & CLAP($\downarrow$) \\ \hline \hline
             & neutral      & 30.71    & 1.020           \\
    Semantic & positive   & 30.67    & 1.056           \\
             & negative     & 30.69    & 1.074           \\ \hline
             \multicolumn{4}{@{}c@{}}{}
\end{tabular}}

\centering \resizebox{0.45\textwidth}{!}{
\begin{tabular}{cc|cc}
    & Affective Word-Output & CLIP($\downarrow$) & CLAP($\downarrow$) \\ \hline \hline
    & positive-neutral     & 37.88    & 1.360           \\
    & positive-positive    & 37.87    & 1.351           \\
    Affect & positive-negative    & 37.84    & 1.368           \\
    Difference & negative-neutral     & 38.41    & 1.270            \\
    & negative-positive    & 38.38    & 1.268           \\
    & negative-negative    & 38.36    & 1.253          \\ \hline
           
\end{tabular}}
\caption{Quantitative results calculating distances between text prompt and output (thumbnail image and music) at CLIP and CLAP spaces.}
\label{tab:quantitative}
\end{table}

\begin{figure}[ht]
    \centering
    \includegraphics[width=0.40\textwidth]{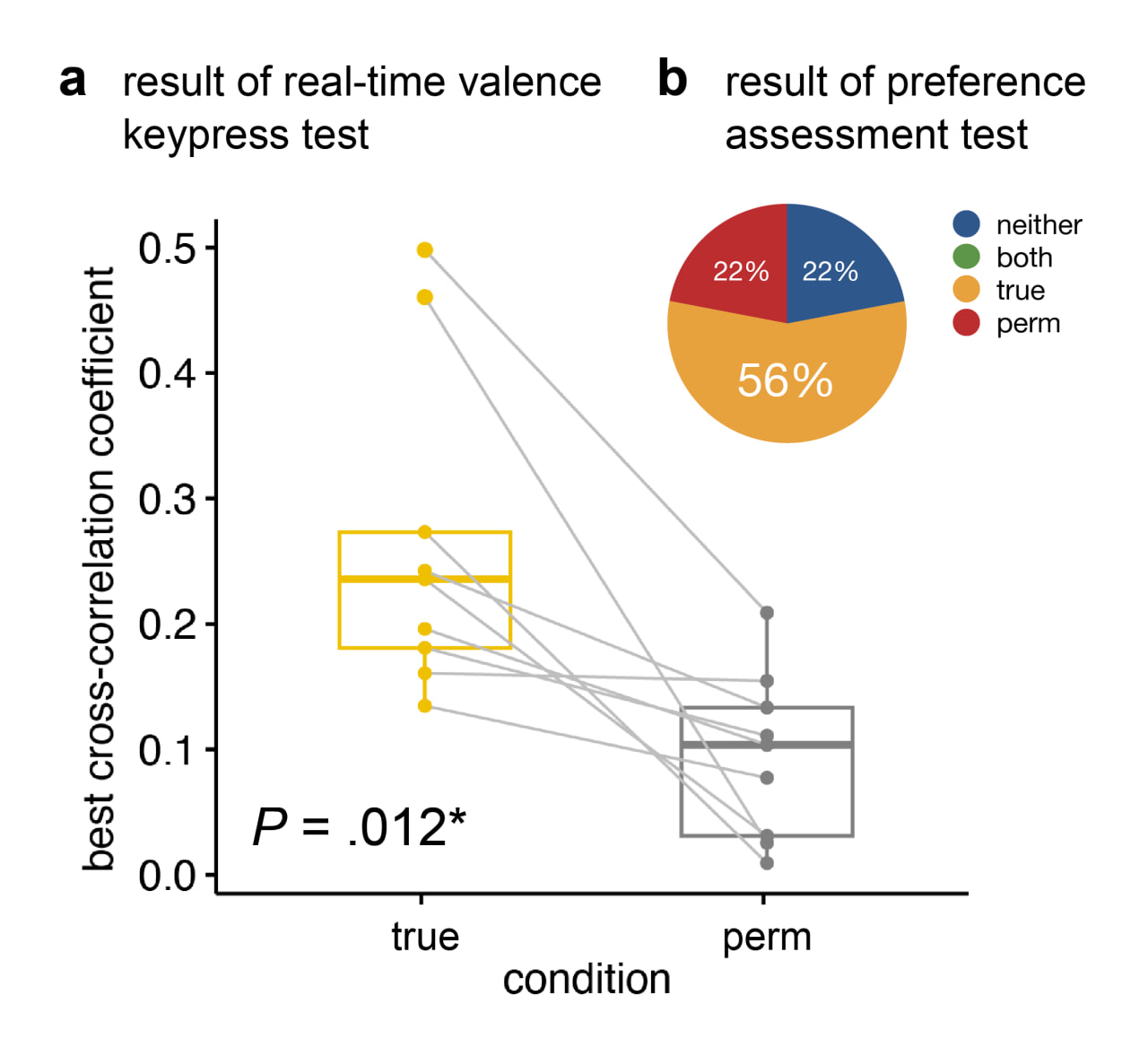}
    \caption{User study results. (a) Cross-correlation coefficients of real-time valence keypresses between the session 1 and 2. (b) Preference assessment results. }
    \label{fig:fig6}
    \Description{.}
\end{figure}

\subsection{Quantitative Evaluation} We show quantitative evaluation in Tab.~\ref{tab:quantitative} and Figure~\ref{fig:fig5}. 
Semantic evaluation using CLIP and CLAP embeddings shows consistent distances between reference prompts (i.e., prompts without affective words) and affect-contextualized outputs across affective states, indicating semantic coherence. 
Additionally, the minimal distances between outputs and their corresponding affective word embeddings (e.g., positive-positive pairs) validate our model's ability to effectively direct content generation toward intended affective states.

We analyzed whether the generated videos exhibited variations in visual (e.g., hue, saturation, value) and musical features (e.g., mode, emotional brightness, intensity) depending on participants' affective states. To this end, we computed eta-squared, the effect size of a one-way ANOVA, to assess whether each audio and visual feature showed significant differences across three affective states for each participant. Results revealed that, except for saturation, all audiovisual features showed larger effect sizes in the true condition compared to the fake condition (Figure 5a), indicating that the videos in the true condition more accurately reflected participants' valence-dependent characteristics. However, within the true condition, we observed considerable individual differences in the effect sizes of affective states across features (Figure 5b). For example, while subject 2's video displayed large changes in musical features based on affective states, subject 6's clip showed notable effect sizes of valence in visual features.

\subsection{User Study}
Participants showed valence dynamics more closely aligned with their valence reports during Session 1 when viewing the CEBRA-decoded true video compared to the fake video (see Figure~\ref{fig:fig6}a). A Wilcoxon rank-sum test on the best cross-correlation coefficients between real-time valence keypresses in Session 1 and 2 revealed that the true condition produced significantly higher coefficients than the perm condition (${\mathbf r}_{true}=0.265$, ${\mathbf r}_{perm}=0.095$,  ${\textit{p}}$ = .012). Additionally, about 56\% of participants reported that the true video better reflected their memories and associated affective changes (see Figure~\ref{fig:fig6}b). In a preference rating, 5 out of 9 participants chose the true video. These results suggest that our approach can be promising to capture one’s idiosyncratic affect dynamics in their personal memories and generate them as music videos. 

\section{Conclusion \& Future Works}
This research makes significant contributions to cognitive/affective neuroscience and BCI. We introduce \textbf{RevisitAffectiveMemory}, a novel task for studying dynamic affective states during autobiographical memory recall, supported by a comprehensive multimodal dataset (\textbf{EEG-AffectiveMemory}). We demonstrate successful temporal affective trajectory decoding using CEBRA and present \textbf{RYM}, a framework generating personalized audiovisual content synchronized with individual affective experiences.

Our evaluation, incorporating real-time user reports, temporal alignment of affective trajectories with generated content attributes (visual and musical), and representation-level validation (CLIP \& CLAP), demonstrates the effectiveness of our approach in faithfully reconstructing affect-contextualized memories.

Our study aimed to provide a proof-of-concept, demonstrating the feasibility of our proposed framework. To this end, we intentionally selected a homogeneous sample (right-handed Korean young adults, ages 22-28, without specific neurological conditions) to control for potential confounding variables such as cultural background, age, and mental wellness. Similarly, we utilized established neural network components for our framework to ensure extensibility and compatibility with various generative models. While these methodological choices focused our assessment on the framework’s capability, they introduced limitations in sample diversity and technical novelty. Nevertheless, our sample size aligns with prior foundational studies in EEG-based visual reconstruction \cite{shimizu2022improving, lee2021subject}, and our approach successfully reconstructs affect-contextualized memories, highlighting the potential of our novel task for AI-driven personal content generation. In future studies, we will test the broader applicability of our framework with larger and more diverse samples to account for the idiosyncratic nature of emotional memory. Also, a more systematic investigation into the interaction between the characteristics of recalled memories and the features of the generated videos is warranted in future research (e.g., \textit{how do generated videos differ between memories of relatively recent events and those from early childhood?}). Last, we will improve technical limitations of our framework, such as the smoothness of transitions between affective states in the generated content. 



\begin{acks}
\sloppy 
This work was supported by the National Research Foundation of Korea (NRF) grant funded by the Korea government (MSIT) (No. 2021R1C1C1006503, RS\allowbreak{}-\allowbreak{}2023-00266787, RS\allowbreak{}-\allowbreak{}2023\allowbreak{}-\allowbreak{}00265406, RS-2024-00421268, RS\allowbreak{}-\allowbreak{}2024\allowbreak{}-\allowbreak{}00342301, RS\allowbreak{}-\allowbreak{}2024-00435727, NRF\allowbreak{}-\allowbreak{}2021M\allowbreak{}3E5D2A0102\allowbreak{}2515), by Creative-Pioneering Researchers Program through Seoul National University (No. 200\allowbreak{}-\allowbreak{}20240057, 200\allowbreak{}-\allowbreak{}20240135), by Semi-Supervised Learning Research Grant by SAMSUNG (No.\allowbreak{}A0342\allowbreak{}-\allowbreak{}20220009), by Identify the network of brain preparation steps for concentration Research Grant by LooxidLabs (No.339\allowbreak{}-\allowbreak{}20230001), by Institute of Information \& communications Technology Planning \& Evaluation (IITP) grant funded by the Korea government(MSIT) [NO.RS\allowbreak{}-\allowbreak{}2021\allowbreak{}-\allowbreak{}II211343, Artificial Intelligence Graduate School Program (Seoul National University)] by the MSIT (Ministry of Science, ICT), Korea, under the Global Research Support Program in the Digital Field program (RS\allowbreak{}-\allowbreak{}2024\allowbreak{}-\allowbreak{}00421268) supervised by the IITP (Institute for Information \& Communications Technology Planning \& Evaluation), by the National Supercomputing Center with supercomputing resources including technical support (KSC\allowbreak{}-\allowbreak{}2023\allowbreak{}-\allowbreak{}CRE\allowbreak{}-\allowbreak{}0568), by the Ministry of Education of the Republic of Korea and the National Research Foundation of Korea (NRF\allowbreak{}-\allowbreak{}2021S1A3A2A02090597), by the Korea Health Industry Development Institute (KHIDI), and by the Ministry of Health and Welfare, Republic of Korea (HR22C1605), by Artificial intelligence industrial convergence cluster development project funded by the Ministry of Science and ICT (MSIT, Korea) \& Gwangju Metropolitan City and by KBRI basic research program  through  Korea  Brain  Research  Institute funded by Ministry of Science and ICT (25\allowbreak{}-\allowbreak{}BR\allowbreak{}-\allowbreak{}05\allowbreak{}-\allowbreak{}01). 

This work was also supported by the U.S. Department of Energy (DOE), Office of Science (SC), Advanced Scientific Computing Research program under award DE-SC-0012704 and used resources of the National Energy Research Scientific Computing Center, a DOE Office of Science User Facility supported by the Office of Science of the U.S. Department of Energy under Contract No. DE-AC02-05CH11231 using NERSC award ASCR-ERCAP0033711.
\end{acks}

\bibliographystyle{ACM-Reference-Format}
\balance
\bibliography{main}


\begin{thebibliography}{55}


\ifx \showCODEN    \undefined \def \showCODEN     #1{\unskip}     \fi
\ifx \showISBNx    \undefined \def \showISBNx     #1{\unskip}     \fi
\ifx \showISBNxiii \undefined \def \showISBNxiii  #1{\unskip}     \fi
\ifx \showISSN     \undefined \def \showISSN      #1{\unskip}     \fi
\ifx \showLCCN     \undefined \def \showLCCN      #1{\unskip}     \fi
\ifx \shownote     \undefined \def \shownote      #1{#1}          \fi
\ifx \showarticletitle \undefined \def \showarticletitle #1{#1}   \fi
\ifx \showURL      \undefined \def \showURL       {\relax}        \fi
\providecommand\bibfield[2]{#2}
\providecommand\bibinfo[2]{#2}
\providecommand\natexlab[1]{#1}
\providecommand\showeprint[2][]{arXiv:#2}

\bibitem[Agostinelli et~al\mbox{.}(2023)]%
        {agostinelli2023musiclm}
\bibfield{author}{\bibinfo{person}{Andrea Agostinelli}, \bibinfo{person}{Timo~I Denk}, \bibinfo{person}{Zal{\'a}n Borsos}, \bibinfo{person}{Jesse Engel}, \bibinfo{person}{Mauro Verzetti}, \bibinfo{person}{Antoine Caillon}, \bibinfo{person}{Qingqing Huang}, \bibinfo{person}{Aren Jansen}, \bibinfo{person}{Adam Roberts}, \bibinfo{person}{Marco Tagliasacchi}, {et~al\mbox{.}}} \bibinfo{year}{2023}\natexlab{}.
\newblock \showarticletitle{Musiclm: Generating music from text}.
\newblock \bibinfo{journal}{\emph{arXiv preprint arXiv:2301.11325}} (\bibinfo{year}{2023}).
\newblock


\bibitem[Al-Nafjan et~al\mbox{.}(2017)]%
        {al2017review}
\bibfield{author}{\bibinfo{person}{Abeer Al-Nafjan}, \bibinfo{person}{Manar Hosny}, \bibinfo{person}{Yousef Al-Ohali}, {and} \bibinfo{person}{Areej Al-Wabil}.} \bibinfo{year}{2017}\natexlab{}.
\newblock \showarticletitle{Review and classification of emotion recognition based on EEG brain-computer interface system research: a systematic review}.
\newblock \bibinfo{journal}{\emph{Applied Sciences}} \bibinfo{volume}{7}, \bibinfo{number}{12} (\bibinfo{year}{2017}), \bibinfo{pages}{1239}.
\newblock


\bibitem[Chandran et~al\mbox{.}(2021)]%
        {chandran2021adaptive}
\bibfield{author}{\bibinfo{person}{Prashanth Chandran}, \bibinfo{person}{Gaspard Zoss}, \bibinfo{person}{Paulo Gotardo}, \bibinfo{person}{Markus Gross}, {and} \bibinfo{person}{Derek Bradley}.} \bibinfo{year}{2021}\natexlab{}.
\newblock \showarticletitle{Adaptive convolutions for structure-aware style transfer}. In \bibinfo{booktitle}{\emph{Proceedings of the IEEE/CVF conference on computer vision and pattern recognition}}. \bibinfo{pages}{7972--7981}.
\newblock


\bibitem[Chanel et~al\mbox{.}(2009)]%
        {chanel2009short}
\bibfield{author}{\bibinfo{person}{Guillaume Chanel}, \bibinfo{person}{Joep~JM Kierkels}, \bibinfo{person}{Mohammad Soleymani}, {and} \bibinfo{person}{Thierry Pun}.} \bibinfo{year}{2009}\natexlab{}.
\newblock \showarticletitle{Short-term emotion assessment in a recall paradigm}.
\newblock \bibinfo{journal}{\emph{International Journal of Human-Computer Studies}} \bibinfo{volume}{67}, \bibinfo{number}{8} (\bibinfo{year}{2009}), \bibinfo{pages}{607--627}.
\newblock


\bibitem[Chen et~al\mbox{.}(2022)]%
        {chen2022emotion}
\bibfield{author}{\bibinfo{person}{Jin Chen}, \bibinfo{person}{Tony Ro}, {and} \bibinfo{person}{Zhigang Zhu}.} \bibinfo{year}{2022}\natexlab{}.
\newblock \showarticletitle{Emotion recognition with audio, video, EEG, and EMG: a dataset and baseline approaches}.
\newblock \bibinfo{journal}{\emph{IEEE Access}}  \bibinfo{volume}{10} (\bibinfo{year}{2022}), \bibinfo{pages}{13229--13242}.
\newblock


\bibitem[Chen and Schmidt(2016)]%
        {chen2016fast}
\bibfield{author}{\bibinfo{person}{Tian~Qi Chen} {and} \bibinfo{person}{Mark Schmidt}.} \bibinfo{year}{2016}\natexlab{}.
\newblock \showarticletitle{Fast patch-based style transfer of arbitrary style}.
\newblock \bibinfo{journal}{\emph{arXiv preprint arXiv:1612.04337}} (\bibinfo{year}{2016}).
\newblock


\bibitem[Chung et~al\mbox{.}(2024)]%
        {chung2024style}
\bibfield{author}{\bibinfo{person}{Jiwoo Chung}, \bibinfo{person}{Sangeek Hyun}, {and} \bibinfo{person}{Jae-Pil Heo}.} \bibinfo{year}{2024}\natexlab{}.
\newblock \showarticletitle{Style injection in diffusion: A training-free approach for adapting large-scale diffusion models for style transfer}. In \bibinfo{booktitle}{\emph{Proceedings of the IEEE/CVF Conference on Computer Vision and Pattern Recognition}}. \bibinfo{pages}{8795--8805}.
\newblock


\bibitem[C{\'\i}fka et~al\mbox{.}(2019)]%
        {cifka2019supervised}
\bibfield{author}{\bibinfo{person}{Ond{\v{r}}ej C{\'\i}fka}, \bibinfo{person}{Umut {\c{S}}im{\c{s}}ekli}, {and} \bibinfo{person}{Ga{\"e}l Richard}.} \bibinfo{year}{2019}\natexlab{}.
\newblock \showarticletitle{Supervised symbolic music style translation using synthetic data}.
\newblock \bibinfo{journal}{\emph{arXiv preprint arXiv:1907.02265}} (\bibinfo{year}{2019}).
\newblock


\bibitem[Copet et~al\mbox{.}(2024)]%
        {copet2024simple}
\bibfield{author}{\bibinfo{person}{Jade Copet}, \bibinfo{person}{Felix Kreuk}, \bibinfo{person}{Itai Gat}, \bibinfo{person}{Tal Remez}, \bibinfo{person}{David Kant}, \bibinfo{person}{Gabriel Synnaeve}, \bibinfo{person}{Yossi Adi}, {and} \bibinfo{person}{Alexandre D{\'e}fossez}.} \bibinfo{year}{2024}\natexlab{}.
\newblock \showarticletitle{Simple and controllable music generation}.
\newblock \bibinfo{journal}{\emph{Advances in Neural Information Processing Systems}}  \bibinfo{volume}{36} (\bibinfo{year}{2024}).
\newblock


\bibitem[Dai et~al\mbox{.}(2018)]%
        {dai2018music}
\bibfield{author}{\bibinfo{person}{Shuqi Dai}, \bibinfo{person}{Zheng Zhang}, {and} \bibinfo{person}{Gus~G Xia}.} \bibinfo{year}{2018}\natexlab{}.
\newblock \showarticletitle{Music style transfer: A position paper}.
\newblock \bibinfo{journal}{\emph{arXiv preprint arXiv:1803.06841}} (\bibinfo{year}{2018}).
\newblock


\bibitem[Dar et~al\mbox{.}(2024)]%
        {dar2024insights}
\bibfield{author}{\bibinfo{person}{Muhammad~Najam Dar}, \bibinfo{person}{Muhammad~Usman Akram}, \bibinfo{person}{Ahmad~Rauf Subhani}, \bibinfo{person}{Sajid~Gul Khawaja}, \bibinfo{person}{Constantino~Carlos Reyes-Aldasoro}, {and} \bibinfo{person}{Sarah Gul}.} \bibinfo{year}{2024}\natexlab{}.
\newblock \showarticletitle{Insights from EEG analysis of evoked memory recalls using deep learning for emotion charting}.
\newblock \bibinfo{journal}{\emph{Scientific Reports}} \bibinfo{volume}{14}, \bibinfo{number}{1} (\bibinfo{year}{2024}), \bibinfo{pages}{17080}.
\newblock


\bibitem[Engel et~al\mbox{.}(2017)]%
        {engel2017neural}
\bibfield{author}{\bibinfo{person}{Jesse Engel}, \bibinfo{person}{Cinjon Resnick}, \bibinfo{person}{Adam Roberts}, \bibinfo{person}{Sander Dieleman}, \bibinfo{person}{Mohammad Norouzi}, \bibinfo{person}{Douglas Eck}, {and} \bibinfo{person}{Karen Simonyan}.} \bibinfo{year}{2017}\natexlab{}.
\newblock \showarticletitle{Neural audio synthesis of musical notes with wavenet autoencoders}. In \bibinfo{booktitle}{\emph{International Conference on Machine Learning}}. PMLR, \bibinfo{pages}{1068--1077}.
\newblock


\bibitem[Everaert et~al\mbox{.}(2023)]%
        {everaert2023diffusion}
\bibfield{author}{\bibinfo{person}{Martin~Nicolas Everaert}, \bibinfo{person}{Marco Bocchio}, \bibinfo{person}{Sami Arpa}, \bibinfo{person}{Sabine S{\"u}sstrunk}, {and} \bibinfo{person}{Radhakrishna Achanta}.} \bibinfo{year}{2023}\natexlab{}.
\newblock \showarticletitle{Diffusion in style}. In \bibinfo{booktitle}{\emph{Proceedings of the IEEE/CVF International Conference on Computer Vision}}. \bibinfo{pages}{2251--2261}.
\newblock


\bibitem[Gatys et~al\mbox{.}(2016)]%
        {gatys2016image}
\bibfield{author}{\bibinfo{person}{Leon~A Gatys}, \bibinfo{person}{Alexander~S Ecker}, {and} \bibinfo{person}{Matthias Bethge}.} \bibinfo{year}{2016}\natexlab{}.
\newblock \showarticletitle{Image style transfer using convolutional neural networks}. In \bibinfo{booktitle}{\emph{Proceedings of the IEEE conference on computer vision and pattern recognition}}. \bibinfo{pages}{2414--2423}.
\newblock


\bibitem[Grinstein et~al\mbox{.}(2018)]%
        {grinstein2018audio}
\bibfield{author}{\bibinfo{person}{Eric Grinstein}, \bibinfo{person}{Ngoc~QK Duong}, \bibinfo{person}{Alexey Ozerov}, {and} \bibinfo{person}{Patrick P{\'e}rez}.} \bibinfo{year}{2018}\natexlab{}.
\newblock \showarticletitle{Audio style transfer}. In \bibinfo{booktitle}{\emph{2018 IEEE international conference on acoustics, speech and signal processing (ICASSP)}}. IEEE, \bibinfo{pages}{586--590}.
\newblock


\bibitem[Huang et~al\mbox{.}(2018b)]%
        {huang2018automatic}
\bibfield{author}{\bibinfo{person}{Jing Huang}, \bibinfo{person}{Shizhe Zhou}, \bibinfo{person}{Xianyi Zhu}, \bibinfo{person}{Yiwen Li}, {and} \bibinfo{person}{Chengfeng Zhou}.} \bibinfo{year}{2018}\natexlab{b}.
\newblock \showarticletitle{Automatic image style transfer using emotion-palette}. In \bibinfo{booktitle}{\emph{Tenth International Conference on Digital Image Processing (ICDIP 2018)}}, Vol.~\bibinfo{volume}{10806}. SPIE, \bibinfo{pages}{1197--1206}.
\newblock


\bibitem[Huang et~al\mbox{.}(2018a)]%
        {huang2018timbretron}
\bibfield{author}{\bibinfo{person}{Sicong Huang}, \bibinfo{person}{Qiyang Li}, \bibinfo{person}{Cem Anil}, \bibinfo{person}{Xuchan Bao}, \bibinfo{person}{Sageev Oore}, {and} \bibinfo{person}{Roger~B Grosse}.} \bibinfo{year}{2018}\natexlab{a}.
\newblock \showarticletitle{Timbretron: A wavenet (cyclegan (cqt (audio))) pipeline for musical timbre transfer}.
\newblock \bibinfo{journal}{\emph{arXiv preprint arXiv:1811.09620}} (\bibinfo{year}{2018}).
\newblock


\bibitem[Huang and Belongie(2017)]%
        {huang2017arbitrary}
\bibfield{author}{\bibinfo{person}{Xun Huang} {and} \bibinfo{person}{Serge Belongie}.} \bibinfo{year}{2017}\natexlab{}.
\newblock \showarticletitle{Arbitrary style transfer in real-time with adaptive instance normalization}. In \bibinfo{booktitle}{\emph{Proceedings of the IEEE international conference on computer vision}}. \bibinfo{pages}{1501--1510}.
\newblock


\bibitem[Iacoviello et~al\mbox{.}(2015)]%
        {iacoviello2015real}
\bibfield{author}{\bibinfo{person}{Daniela Iacoviello}, \bibinfo{person}{Andrea Petracca}, \bibinfo{person}{Matteo Spezialetti}, {and} \bibinfo{person}{Giuseppe Placidi}.} \bibinfo{year}{2015}\natexlab{}.
\newblock \showarticletitle{A real-time classification algorithm for EEG-based BCI driven by self-induced emotions}.
\newblock \bibinfo{journal}{\emph{Computer methods and programs in biomedicine}} \bibinfo{volume}{122}, \bibinfo{number}{3} (\bibinfo{year}{2015}), \bibinfo{pages}{293--303}.
\newblock


\bibitem[J{\"a}ncke(2008)]%
        {jancke2008music}
\bibfield{author}{\bibinfo{person}{Lutz J{\"a}ncke}.} \bibinfo{year}{2008}\natexlab{}.
\newblock \showarticletitle{Music, memory and emotion}.
\newblock \bibinfo{journal}{\emph{Journal of biology}}  \bibinfo{volume}{7} (\bibinfo{year}{2008}), \bibinfo{pages}{1--5}.
\newblock


\bibitem[Johnson et~al\mbox{.}(2016)]%
        {johnson2016perceptual}
\bibfield{author}{\bibinfo{person}{Justin Johnson}, \bibinfo{person}{Alexandre Alahi}, {and} \bibinfo{person}{Li Fei-Fei}.} \bibinfo{year}{2016}\natexlab{}.
\newblock \showarticletitle{Perceptual losses for real-time style transfer and super-resolution}. In \bibinfo{booktitle}{\emph{Computer Vision--ECCV 2016: 14th European Conference, Amsterdam, The Netherlands, October 11-14, 2016, Proceedings, Part II 14}}. Springer, \bibinfo{pages}{694--711}.
\newblock


\bibitem[Jolly et~al\mbox{.}(2022)]%
        {jolly2022recovering}
\bibfield{author}{\bibinfo{person}{Eshin Jolly}, \bibinfo{person}{Max Farrens}, \bibinfo{person}{Nathan Greenstein}, \bibinfo{person}{Hedwig Eisenbarth}, \bibinfo{person}{Marianne~C Reddan}, \bibinfo{person}{Eric Andrews}, \bibinfo{person}{Tor~D Wager}, {and} \bibinfo{person}{Luke~J Chang}.} \bibinfo{year}{2022}\natexlab{}.
\newblock \showarticletitle{Recovering individual emotional states from sparse ratings using collaborative filtering}.
\newblock \bibinfo{journal}{\emph{Affective Science}} \bibinfo{volume}{3}, \bibinfo{number}{4} (\bibinfo{year}{2022}), \bibinfo{pages}{799--817}.
\newblock


\bibitem[Katsigiannis and Ramzan(2017)]%
        {katsigiannis2017dreamer}
\bibfield{author}{\bibinfo{person}{Stamos Katsigiannis} {and} \bibinfo{person}{Naeem Ramzan}.} \bibinfo{year}{2017}\natexlab{}.
\newblock \showarticletitle{DREAMER: A database for emotion recognition through EEG and ECG signals from wireless low-cost off-the-shelf devices}.
\newblock \bibinfo{journal}{\emph{IEEE journal of biomedical and health informatics}} \bibinfo{volume}{22}, \bibinfo{number}{1} (\bibinfo{year}{2017}), \bibinfo{pages}{98--107}.
\newblock


\bibitem[Kim et~al\mbox{.}(2024)]%
        {kim2024training}
\bibfield{author}{\bibinfo{person}{Sooyoung Kim}, \bibinfo{person}{Joonwoo Kwon}, \bibinfo{person}{Heehwan Wang}, \bibinfo{person}{Shinjae Yoo}, \bibinfo{person}{Yuewei Lin}, {and} \bibinfo{person}{Jiook Cha}.} \bibinfo{year}{2024}\natexlab{}.
\newblock \showarticletitle{A Training-Free Approach for Music Style Transfer with Latent Diffusion Models}.
\newblock \bibinfo{journal}{\emph{arXiv preprint arXiv:2411.15913}} (\bibinfo{year}{2024}).
\newblock


\bibitem[Koelstra et~al\mbox{.}(2011)]%
        {koelstra2011deap}
\bibfield{author}{\bibinfo{person}{Sander Koelstra}, \bibinfo{person}{Christian Muhl}, \bibinfo{person}{Mohammad Soleymani}, \bibinfo{person}{Jong-Seok Lee}, \bibinfo{person}{Ashkan Yazdani}, \bibinfo{person}{Touradj Ebrahimi}, \bibinfo{person}{Thierry Pun}, \bibinfo{person}{Anton Nijholt}, {and} \bibinfo{person}{Ioannis Patras}.} \bibinfo{year}{2011}\natexlab{}.
\newblock \showarticletitle{Deap: A database for emotion analysis; using physiological signals}.
\newblock \bibinfo{journal}{\emph{IEEE transactions on affective computing}} \bibinfo{volume}{3}, \bibinfo{number}{1} (\bibinfo{year}{2011}), \bibinfo{pages}{18--31}.
\newblock


\bibitem[Kotovenko et~al\mbox{.}(2019)]%
        {kotovenko2019content}
\bibfield{author}{\bibinfo{person}{Dmytro Kotovenko}, \bibinfo{person}{Artsiom Sanakoyeu}, \bibinfo{person}{Sabine Lang}, {and} \bibinfo{person}{Bjorn Ommer}.} \bibinfo{year}{2019}\natexlab{}.
\newblock \showarticletitle{Content and style disentanglement for artistic style transfer}. In \bibinfo{booktitle}{\emph{Proceedings of the IEEE/CVF international conference on computer vision}}. \bibinfo{pages}{4422--4431}.
\newblock


\bibitem[Kuppens and Verduyn(2017)]%
        {kuppens2017emotion}
\bibfield{author}{\bibinfo{person}{Peter Kuppens} {and} \bibinfo{person}{Philippe Verduyn}.} \bibinfo{year}{2017}\natexlab{}.
\newblock \showarticletitle{Emotion dynamics}.
\newblock \bibinfo{journal}{\emph{Current Opinion in Psychology}}  \bibinfo{volume}{17} (\bibinfo{year}{2017}), \bibinfo{pages}{22--26}.
\newblock


\bibitem[Kwon and Ye(2022)]%
        {kwon2022clipstyler}
\bibfield{author}{\bibinfo{person}{Gihyun Kwon} {and} \bibinfo{person}{Jong~Chul Ye}.} \bibinfo{year}{2022}\natexlab{}.
\newblock \showarticletitle{Clipstyler: Image style transfer with a single text condition}. In \bibinfo{booktitle}{\emph{Proceedings of the IEEE/CVF Conference on Computer Vision and Pattern Recognition}}. \bibinfo{pages}{18062--18071}.
\newblock


\bibitem[Kwon et~al\mbox{.}(2024)]%
        {kwon2024aesfa}
\bibfield{author}{\bibinfo{person}{Joonwoo Kwon}, \bibinfo{person}{Sooyoung Kim}, \bibinfo{person}{Yuewei Lin}, \bibinfo{person}{Shinjae Yoo}, {and} \bibinfo{person}{Jiook Cha}.} \bibinfo{year}{2024}\natexlab{}.
\newblock \showarticletitle{Aesfa: an aesthetic feature-aware arbitrary neural style transfer}. In \bibinfo{booktitle}{\emph{Proceedings of the AAAI conference on artificial intelligence}}, Vol.~\bibinfo{volume}{38}. \bibinfo{pages}{13310--13319}.
\newblock


\bibitem[Lee et~al\mbox{.}(2021)]%
        {lee2021subject}
\bibfield{author}{\bibinfo{person}{Pilhyeon Lee}, \bibinfo{person}{Sunhee Hwang}, \bibinfo{person}{Seogkyu Jeon}, {and} \bibinfo{person}{Hyeran Byun}.} \bibinfo{year}{2021}\natexlab{}.
\newblock \showarticletitle{Subject adaptive eeg-based visual recognition}. In \bibinfo{booktitle}{\emph{Asian Conference on Pattern Recognition}}. Springer, \bibinfo{pages}{322--334}.
\newblock


\bibitem[Liu et~al\mbox{.}(2025)]%
        {liu2025javisdit}
\bibfield{author}{\bibinfo{person}{Kai Liu}, \bibinfo{person}{Wei Li}, \bibinfo{person}{Lai Chen}, \bibinfo{person}{Shengqiong Wu}, \bibinfo{person}{Yanhao Zheng}, \bibinfo{person}{Jiayi Ji}, \bibinfo{person}{Fan Zhou}, \bibinfo{person}{Rongxin Jiang}, \bibinfo{person}{Jiebo Luo}, \bibinfo{person}{Hao Fei}, {et~al\mbox{.}}} \bibinfo{year}{2025}\natexlab{}.
\newblock \showarticletitle{Javisdit: Joint audio-video diffusion transformer with hierarchical spatio-temporal prior synchronization}.
\newblock \bibinfo{journal}{\emph{arXiv preprint arXiv:2503.23377}} (\bibinfo{year}{2025}).
\newblock


\bibitem[McClay et~al\mbox{.}(2023)]%
        {mcclay2023dynamic}
\bibfield{author}{\bibinfo{person}{Mason McClay}, \bibinfo{person}{Matthew~E Sachs}, {and} \bibinfo{person}{David Clewett}.} \bibinfo{year}{2023}\natexlab{}.
\newblock \showarticletitle{Dynamic emotional states shape the episodic structure of memory}.
\newblock \bibinfo{journal}{\emph{Nature Communications}} \bibinfo{volume}{14}, \bibinfo{number}{1} (\bibinfo{year}{2023}), \bibinfo{pages}{6533}.
\newblock


\bibitem[Merk et~al\mbox{.}(2023)]%
        {merk2023invasive}
\bibfield{author}{\bibinfo{person}{Timon Merk}, \bibinfo{person}{Richard K{\"o}hler}, \bibinfo{person}{Victoria Peterson}, \bibinfo{person}{Laura Lyra}, \bibinfo{person}{Jonathan Vanhoecke}, \bibinfo{person}{Meera Chikermane}, \bibinfo{person}{Thomas Binns}, \bibinfo{person}{Ningfei Li}, \bibinfo{person}{Ashley Walton}, \bibinfo{person}{Alan Bush}, {et~al\mbox{.}}} \bibinfo{year}{2023}\natexlab{}.
\newblock \showarticletitle{Invasive neurophysiology and whole brain connectomics for neural decoding in patients with brain implants}.
\newblock \bibinfo{journal}{\emph{Research Square}} (\bibinfo{year}{2023}).
\newblock


\bibitem[Mills and D'Mello(2014)]%
        {mills2014validity}
\bibfield{author}{\bibinfo{person}{Caitlin Mills} {and} \bibinfo{person}{Sidney D'Mello}.} \bibinfo{year}{2014}\natexlab{}.
\newblock \showarticletitle{On the validity of the autobiographical emotional memory task for emotion induction}.
\newblock \bibinfo{journal}{\emph{PloS one}} \bibinfo{volume}{9}, \bibinfo{number}{4} (\bibinfo{year}{2014}), \bibinfo{pages}{e95837}.
\newblock


\bibitem[Miranda-Correa et~al\mbox{.}(2018)]%
        {miranda2018amigos}
\bibfield{author}{\bibinfo{person}{Juan~Abdon Miranda-Correa}, \bibinfo{person}{Mojtaba~Khomami Abadi}, \bibinfo{person}{Nicu Sebe}, {and} \bibinfo{person}{Ioannis Patras}.} \bibinfo{year}{2018}\natexlab{}.
\newblock \showarticletitle{Amigos: A dataset for affect, personality and mood research on individuals and groups}.
\newblock \bibinfo{journal}{\emph{IEEE transactions on affective computing}} \bibinfo{volume}{12}, \bibinfo{number}{2} (\bibinfo{year}{2018}), \bibinfo{pages}{479--493}.
\newblock


\bibitem[Park and Min(2005)]%
        {park2005making}
\bibfield{author}{\bibinfo{person}{In-Jo Park} {and} \bibinfo{person}{Kyunghwan Min}.} \bibinfo{year}{2005}\natexlab{}.
\newblock \showarticletitle{Making a list of Korean emotion terms and exploring dimensions underlying them}.
\newblock \bibinfo{journal}{\emph{Korean Journal of Social and Personality Psychology}} \bibinfo{volume}{19}, \bibinfo{number}{1} (\bibinfo{year}{2005}), \bibinfo{pages}{109--129}.
\newblock


\bibitem[Paszke et~al\mbox{.}(2019)]%
        {paszke2019pytorch}
\bibfield{author}{\bibinfo{person}{Adam Paszke}, \bibinfo{person}{Sam Gross}, \bibinfo{person}{Francisco Massa}, \bibinfo{person}{Adam Lerer}, \bibinfo{person}{James Bradbury}, \bibinfo{person}{Gregory Chanan}, \bibinfo{person}{Trevor Killeen}, \bibinfo{person}{Zeming Lin}, \bibinfo{person}{Natalia Gimelshein}, \bibinfo{person}{Luca Antiga}, {et~al\mbox{.}}} \bibinfo{year}{2019}\natexlab{}.
\newblock \showarticletitle{Pytorch: An imperative style, high-performance deep learning library}.
\newblock \bibinfo{journal}{\emph{Advances in neural information processing systems}}  \bibinfo{volume}{32} (\bibinfo{year}{2019}).
\newblock


\bibitem[Rombach et~al\mbox{.}(2022)]%
        {rombach2022high}
\bibfield{author}{\bibinfo{person}{Robin Rombach}, \bibinfo{person}{Andreas Blattmann}, \bibinfo{person}{Dominik Lorenz}, \bibinfo{person}{Patrick Esser}, {and} \bibinfo{person}{Bj{\"o}rn Ommer}.} \bibinfo{year}{2022}\natexlab{}.
\newblock \showarticletitle{High-resolution image synthesis with latent diffusion models}. In \bibinfo{booktitle}{\emph{Proceedings of the IEEE/CVF conference on computer vision and pattern recognition}}. \bibinfo{pages}{10684--10695}.
\newblock


\bibitem[Ruan et~al\mbox{.}(2023)]%
        {ruan2023mm}
\bibfield{author}{\bibinfo{person}{Ludan Ruan}, \bibinfo{person}{Yiyang Ma}, \bibinfo{person}{Huan Yang}, \bibinfo{person}{Huiguo He}, \bibinfo{person}{Bei Liu}, \bibinfo{person}{Jianlong Fu}, \bibinfo{person}{Nicholas~Jing Yuan}, \bibinfo{person}{Qin Jin}, {and} \bibinfo{person}{Baining Guo}.} \bibinfo{year}{2023}\natexlab{}.
\newblock \showarticletitle{Mm-diffusion: Learning multi-modal diffusion models for joint audio and video generation}. In \bibinfo{booktitle}{\emph{Proceedings of the IEEE/CVF Conference on Computer Vision and Pattern Recognition}}. \bibinfo{pages}{10219--10228}.
\newblock


\bibitem[Saarim{\"a}ki(2021)]%
        {saarimaki2021naturalistic}
\bibfield{author}{\bibinfo{person}{Heini Saarim{\"a}ki}.} \bibinfo{year}{2021}\natexlab{}.
\newblock \showarticletitle{Naturalistic stimuli in affective neuroimaging: A review}.
\newblock \bibinfo{journal}{\emph{Frontiers in human neuroscience}}  \bibinfo{volume}{15} (\bibinfo{year}{2021}), \bibinfo{pages}{675068}.
\newblock


\bibitem[Schneider et~al\mbox{.}(2023)]%
        {schneider2023learnable}
\bibfield{author}{\bibinfo{person}{Steffen Schneider}, \bibinfo{person}{Jin~Hwa Lee}, {and} \bibinfo{person}{Mackenzie~Weygandt Mathis}.} \bibinfo{year}{2023}\natexlab{}.
\newblock \showarticletitle{Learnable latent embeddings for joint behavioural and neural analysis}.
\newblock \bibinfo{journal}{\emph{Nature}} \bibinfo{volume}{617}, \bibinfo{number}{7960} (\bibinfo{year}{2023}), \bibinfo{pages}{360--368}.
\newblock


\bibitem[Shimizu and Srinivasan(2022)]%
        {shimizu2022improving}
\bibfield{author}{\bibinfo{person}{Hirokatsu Shimizu} {and} \bibinfo{person}{Ramesh Srinivasan}.} \bibinfo{year}{2022}\natexlab{}.
\newblock \showarticletitle{Improving classification and reconstruction of imagined images from EEG signals}.
\newblock \bibinfo{journal}{\emph{Plos one}} \bibinfo{volume}{17}, \bibinfo{number}{9} (\bibinfo{year}{2022}), \bibinfo{pages}{e0274847}.
\newblock


\bibitem[Siedlecka and Denson(2019)]%
        {siedlecka2019experimental}
\bibfield{author}{\bibinfo{person}{Ewa Siedlecka} {and} \bibinfo{person}{Thomas~F Denson}.} \bibinfo{year}{2019}\natexlab{}.
\newblock \showarticletitle{Experimental methods for inducing basic emotions: A qualitative review}.
\newblock \bibinfo{journal}{\emph{Emotion Review}} \bibinfo{volume}{11}, \bibinfo{number}{1} (\bibinfo{year}{2019}), \bibinfo{pages}{87--97}.
\newblock


\bibitem[Sigel et~al\mbox{.}(2021)]%
        {sigel2021music}
\bibfield{author}{\bibinfo{person}{Miles Sigel}, \bibinfo{person}{Michael Zhou}, {and} \bibinfo{person}{Jiebo Luo}.} \bibinfo{year}{2021}\natexlab{}.
\newblock \showarticletitle{Music Sentiment Transfer}.
\newblock \bibinfo{journal}{\emph{arXiv preprint arXiv:2110.05765}} (\bibinfo{year}{2021}).
\newblock


\bibitem[Sun et~al\mbox{.}(2024)]%
        {sun2024mm}
\bibfield{author}{\bibinfo{person}{Mingzhen Sun}, \bibinfo{person}{Weining Wang}, \bibinfo{person}{Yanyuan Qiao}, \bibinfo{person}{Jiahui Sun}, \bibinfo{person}{Zihan Qin}, \bibinfo{person}{Longteng Guo}, \bibinfo{person}{Xinxin Zhu}, {and} \bibinfo{person}{Jing Liu}.} \bibinfo{year}{2024}\natexlab{}.
\newblock \showarticletitle{Mm-ldm: Multi-modal latent diffusion model for sounding video generation}. In \bibinfo{booktitle}{\emph{Proceedings of the 32nd ACM International Conference on Multimedia}}. \bibinfo{pages}{10853--10861}.
\newblock


\bibitem[Tan and Herremans(2020)]%
        {tan2020music}
\bibfield{author}{\bibinfo{person}{Hao~Hao Tan} {and} \bibinfo{person}{Dorien Herremans}.} \bibinfo{year}{2020}\natexlab{}.
\newblock \showarticletitle{Music fadernets: Controllable music generation based on high-level features via low-level feature modelling}.
\newblock \bibinfo{journal}{\emph{arXiv preprint arXiv:2007.15474}} (\bibinfo{year}{2020}).
\newblock


\bibitem[Ulyanov et~al\mbox{.}(2016)]%
        {ulyanov2016texture}
\bibfield{author}{\bibinfo{person}{Dmitry Ulyanov}, \bibinfo{person}{Vadim Lebedev}, \bibinfo{person}{Andrea Vedaldi}, {and} \bibinfo{person}{Victor Lempitsky}.} \bibinfo{year}{2016}\natexlab{}.
\newblock \showarticletitle{Texture networks: Feed-forward synthesis of textures and stylized images}.
\newblock \bibinfo{journal}{\emph{arXiv preprint arXiv:1603.03417}} (\bibinfo{year}{2016}).
\newblock


\bibitem[Vaccaro et~al\mbox{.}(2024)]%
        {vaccaro2024neural}
\bibfield{author}{\bibinfo{person}{Anthony~G Vaccaro}, \bibinfo{person}{Helen Wu}, \bibinfo{person}{Rishab Iyer}, \bibinfo{person}{Shruti Shakthivel}, \bibinfo{person}{Nina~C Christie}, \bibinfo{person}{Antonio Damasio}, {and} \bibinfo{person}{Jonas Kaplan}.} \bibinfo{year}{2024}\natexlab{}.
\newblock \showarticletitle{Neural patterns associated with mixed valence feelings differ in consistency and predictability throughout the brain}.
\newblock \bibinfo{journal}{\emph{Cerebral Cortex}} \bibinfo{volume}{34}, \bibinfo{number}{4} (\bibinfo{year}{2024}), \bibinfo{pages}{bhae122}.
\newblock


\bibitem[Vempati and Sharma(2023)]%
        {vempati2023systematic}
\bibfield{author}{\bibinfo{person}{R Vempati} {and} \bibinfo{person}{L Sharma}.} \bibinfo{year}{2023}\natexlab{}.
\newblock \bibinfo{title}{A systematic review on automated human emotion recognition using electroencephalogram signals and artificial intelligence. Results Eng. 18, 101027 (2023)}.
\newblock


\bibitem[Wang et~al\mbox{.}(2023a)]%
        {wang2023stylediffusion}
\bibfield{author}{\bibinfo{person}{Zhizhong Wang}, \bibinfo{person}{Lei Zhao}, {and} \bibinfo{person}{Wei Xing}.} \bibinfo{year}{2023}\natexlab{a}.
\newblock \showarticletitle{Stylediffusion: Controllable disentangled style transfer via diffusion models}. In \bibinfo{booktitle}{\emph{Proceedings of the IEEE/CVF International Conference on Computer Vision}}. \bibinfo{pages}{7677--7689}.
\newblock


\bibitem[Wang et~al\mbox{.}(2023b)]%
        {wang2023microast}
\bibfield{author}{\bibinfo{person}{Zhizhong Wang}, \bibinfo{person}{Lei Zhao}, \bibinfo{person}{Zhiwen Zuo}, \bibinfo{person}{Ailin Li}, \bibinfo{person}{Haibo Chen}, \bibinfo{person}{Wei Xing}, {and} \bibinfo{person}{Dongming Lu}.} \bibinfo{year}{2023}\natexlab{b}.
\newblock \showarticletitle{MicroAST: Towards Super-Fast Ultra-Resolution Arbitrary Style Transfer}. In \bibinfo{booktitle}{\emph{Proceedings of the AAAI Conference on Artificial Intelligence}}.
\newblock


\bibitem[Weng et~al\mbox{.}(2023)]%
        {weng2023affective}
\bibfield{author}{\bibinfo{person}{Shuchen Weng}, \bibinfo{person}{Peixuan Zhang}, \bibinfo{person}{Zheng Chang}, \bibinfo{person}{Xinlong Wang}, \bibinfo{person}{Si Li}, {and} \bibinfo{person}{Boxin Shi}.} \bibinfo{year}{2023}\natexlab{}.
\newblock \showarticletitle{Affective image filter: Reflecting emotions from text to images}. In \bibinfo{booktitle}{\emph{Proceedings of the IEEE/CVF International Conference on Computer Vision}}. \bibinfo{pages}{10810--10819}.
\newblock


\bibitem[Yi et~al\mbox{.}(2024)]%
        {yi2024awe}
\bibfield{author}{\bibinfo{person}{Jinwoo Yi}, \bibinfo{person}{Danny~Dongyeop Han}, \bibinfo{person}{Seung-Yeop Oh}, {and} \bibinfo{person}{Jiook Cha}.} \bibinfo{year}{2024}\natexlab{}.
\newblock \showarticletitle{Awe is characterized as an ambivalent experience in the human behavior and cortex: integrated virtual reality-electroencephalogram study}.
\newblock \bibinfo{journal}{\emph{bioRxiv}} (\bibinfo{year}{2024}), \bibinfo{pages}{2024--08}.
\newblock


\bibitem[Zhang et~al\mbox{.}(2023)]%
        {zhang2023inversion}
\bibfield{author}{\bibinfo{person}{Yuxin Zhang}, \bibinfo{person}{Nisha Huang}, \bibinfo{person}{Fan Tang}, \bibinfo{person}{Haibin Huang}, \bibinfo{person}{Chongyang Ma}, \bibinfo{person}{Weiming Dong}, {and} \bibinfo{person}{Changsheng Xu}.} \bibinfo{year}{2023}\natexlab{}.
\newblock \showarticletitle{Inversion-based style transfer with diffusion models}. In \bibinfo{booktitle}{\emph{Proceedings of the IEEE/CVF conference on computer vision and pattern recognition}}. \bibinfo{pages}{10146--10156}.
\newblock


\bibitem[Zheng and Lu(2015)]%
        {zheng2015investigating}
\bibfield{author}{\bibinfo{person}{Wei-Long Zheng} {and} \bibinfo{person}{Bao-Liang Lu}.} \bibinfo{year}{2015}\natexlab{}.
\newblock \showarticletitle{Investigating critical frequency bands and channels for EEG-based emotion recognition with deep neural networks}.
\newblock \bibinfo{journal}{\emph{IEEE Transactions on autonomous mental development}} \bibinfo{volume}{7}, \bibinfo{number}{3} (\bibinfo{year}{2015}), \bibinfo{pages}{162--175}.
\newblock


\end{thebibliography}
\end{document}